# PolySet: Restoring the Statistical Ensemble Nature of Polymers for Machine Learning

Khalid FERJI[1]

LCPM, Université de Lorraine, UMR 7375 Nancy – France

Khalid.ferji@univ-lorraine.fr

https://orcid.org/0000-0003-3073-9722

## Abstract

Machine-learning (ML) models in polymer science typically treat a polymer as a single, perfectly defined molecular graph, even though real materials consist of stochastic ensembles of chains with distributed lengths. This mismatch between physical reality and digital representation limits the ability of current models to capture polymer behaviour. Here we introduce PolySet, a framework that represents a polymer as a finite, weighted ensemble of chains sampled from an assumed molar-mass distribution. This ensemble-based encoding is independent of chemical detail, compatible with any molecular representation and illustrated here in the homopolymer case using a minimal language model. We show that PolySet retains higher-order distributional moments (such as $M_z$, $M_{z+1}$), enabling ML models to learn tail-sensitive properties with greatly improved stability and accuracy. By explicitly acknowledging the statistical nature of polymer matter, PolySet establishes a physically grounded foundation for future polymer machine learning, naturally extensible to copolymers, block architectures, and other complex topologies.

## Introduction

Macromolecules[1] derive their identity from both their chemistry and their intrinsic heterogeneity. Every polymer is born as a distribution of chains, and its properties[2-7] emerge from that distribution. This ensemble perspective—established over a century of physical chemistry and macromolecular theory[3,7-9]—remains absent from the representations used in modern machine-learning (ML) models.[10-13] As a result, current artificial intelligent (AI) methods operate with chemically precise yet physically incomplete descriptions of polymeric matter, overlooking the very distribution that governs material behaviour.[14-16]

Although molecular-weight distribution (MWD) is central to polymer physics,[17] it is absent from the structural representations used in polymer informatics. Representations such as SMILES,[18] BigSMILES,[19,20] and SELFIES[21] describe the chemistry of a nominal repeat unit, but they implicitly collapse the polymer to a single "average" molecule. Even when number-average molar mass ($M_n$), weight-average molar mass ($M_w$) and dispersity (Đ) are added as scalar metadata, the latent representation processed by the model remains insufficient.[22,23] For instance, for the rheological properties that depend critically on the high-molecular-weight tail—such as melt viscosity (η)—current ML models remain fundamentally monodisperse in their representation. The recent physics-enforced PENN model[24] uses a single Polymer-Genome fingerprint together with $M_w$ and a scalar Đ, with no explicit encoding of the MWD, despite acknowledging that this latter strongly affects η. Consequently, ML algorithms are trained on chemically correct[25] but physically incomplete inputs, learning from a hypothetical chain rather than from the stochastic ensemble that constitutes the actual material.

To address this foundational limitation, we introduce PolySet, a distribution-aware representation that treats a polymer as a finite stochastic ensemble of chains sampled from its underlying molar-mass distribution (**Fig 1**). Each sampled chain is independently embedded, and the polymer is represented as the probability-weighted aggregation of these embeddings. This construction restores, within ML, the ensemble ontology that has defined polymer physics for more than a century while remaining fully compatible with existing cheminformatics tools.

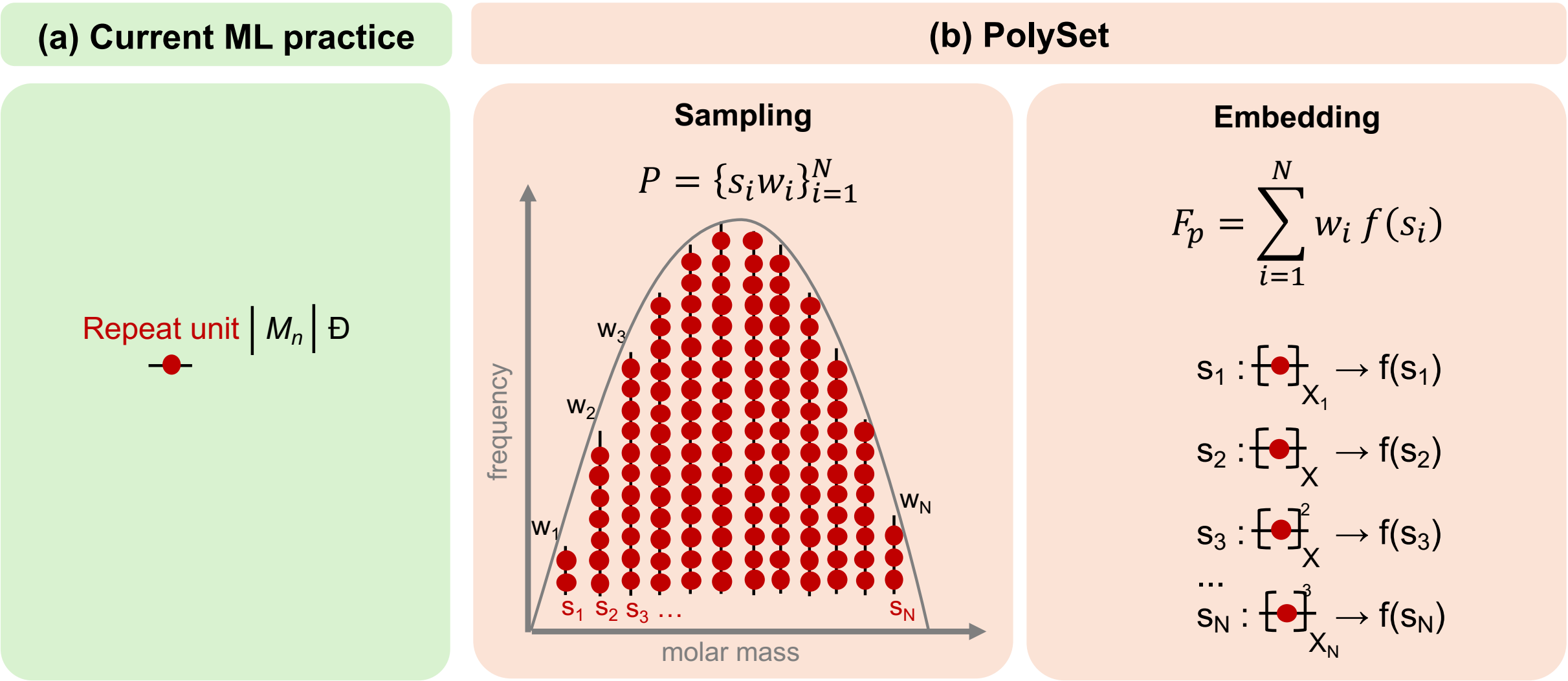

***Fig 1- Conceptual contrast between current ML polymer representations and the PolySet framework.*** *(a) Current ML practice: polymers reduced to a repeat unit and two scalar descriptors. (b) PolySet: polymers represented as a stochastic ensemble of chain lengths with associated weights.*

Importantly, PolySet is not a chemical language and does not replace structural encoders. Rather, it complements them, chemical encoders describe the monomer identity and connectivity, whereas PolySet describes the distribution of chain lengths—the second dimension of polymer identity. A complete polymer representation therefore requires the combination of both structure-aware and distribution-aware descriptors.

In this work, we establish the PolySet framework, demonstrate that finite ensembles faithfully reproduce the target distribution defined by ($M_n$, Đ), and show that distribution-aware embeddings significantly improve model stability and predictive performance. The approach provides a physically grounded and extensible foundation for future ensemble-aware modelling of copolymers, block architectures, and architecturally complex macromolecules.

In the following, we formalize PolySet in the homopolymer case, where the underlying MWD is fully determined by the chain-length distribution. This provides the minimal setting needed to demonstrate how a distribution-aware representation fundamentally extends beyond traditional scalar descriptors. Because the high-molecular-weight tail governs many polymer properties, higher-order moments such as $M_z$ and $M_{z+1}$ provide sensitive probes of distributional structure. PolySet naturally preserves these moments. A lightweight Python package is provided to allow users to generate PolySet embedding.

## Results

Problem statement: degeneracy of $M_n$ & Đ

A fundamental limitation of current polymer informatics arises from the fact that the pair ($M_n$, Đ) does not uniquely determine a polymer's MWD. In practice, polymers synthesized under different kinetic regimes, living versus uncontrolled mechanisms, or distinct processing histories may display identical $M_n$ and Đ while possessing markedly different distribution shapes. These differences may manifest as shifts in modal mass, changes in skewness, or substantial variations in the high-molecular-weight tail—the region that most strongly influences rheology, entanglement density, diffusion, and viscoelastic relaxation.[3,7] Yet none of this structural heterogeneity is represented in current databases, such as PoLyInfo,[26] which typically store only ($M_n$, Đ) as scalar descriptors, nor in the ML encodings derived from them. This degeneracy becomes particularly visible when analyzing higher-order moments of the distribution. Therefore, polymers that are physically distinct become mathematically indistinguishable to learning algorithms, which are unable to resolve differences originating from the distribution itself.

Construction of the synthetic dataset

To quantify the ambiguity introduced by scalar descriptors and to provide a controlled testing ground for distribution-aware representations, we generated an artificial dataset of 10 000 homopolymers spanning realistic ranges of $M_n$ and Đ. The values of $M_n$ were sampled log-uniformly between $10^4$ and $10^6$ g mol$^{-1}$, while dispersities covered the interval $1.5 \leq Đ \leq 4$. **Figs 2a and 2b** show that both quantities are broadly and uniformly represented across the dataset, with no systematic bias toward specific molecular-weight scales or dispersity regimes. The joint distribution of ($M_n$, Đ) pairs, visualized in **Fig 2c**, confirms that the dataset densely covers the two-dimensional space typically encountered in synthetic polymer chemistry.

A key feature of this dataset is the deliberate presence of redundancy: several polymers share the same ($M_n$, Đ) while exhibiting distinct MWD. To quantify these differences, we use higher-order moments of the MWD, including $M_z$, and $M_{z+1}$, which are defined as ratios of progressively weighted moments of the distribution. These moments increasingly emphasize the heavy tail, with $M_{z+1}$ particularly sensitive to rare but massive chains.

Here, we focus on the $M_{z+1}$ moment, which has long been recognized as particularly sensitive to tail variations. As shown in **Fig 2d**, polymers sharing identical ($M_n$, Đ) nevertheless exhibit substantial vertical scatter in $M_{z+1}$, revealing strong degeneracy of the MWD even when the first two moments coincide. Complementary analyses of $M_z$, exploring adjacent orders of tail sensitivity, are provided in **Fig S1**. Together, this moment highlights the intrinsic degeneracy of real polymer databases, where physically distinct distributions collapse onto identical ($M_n$, Đ) entries.

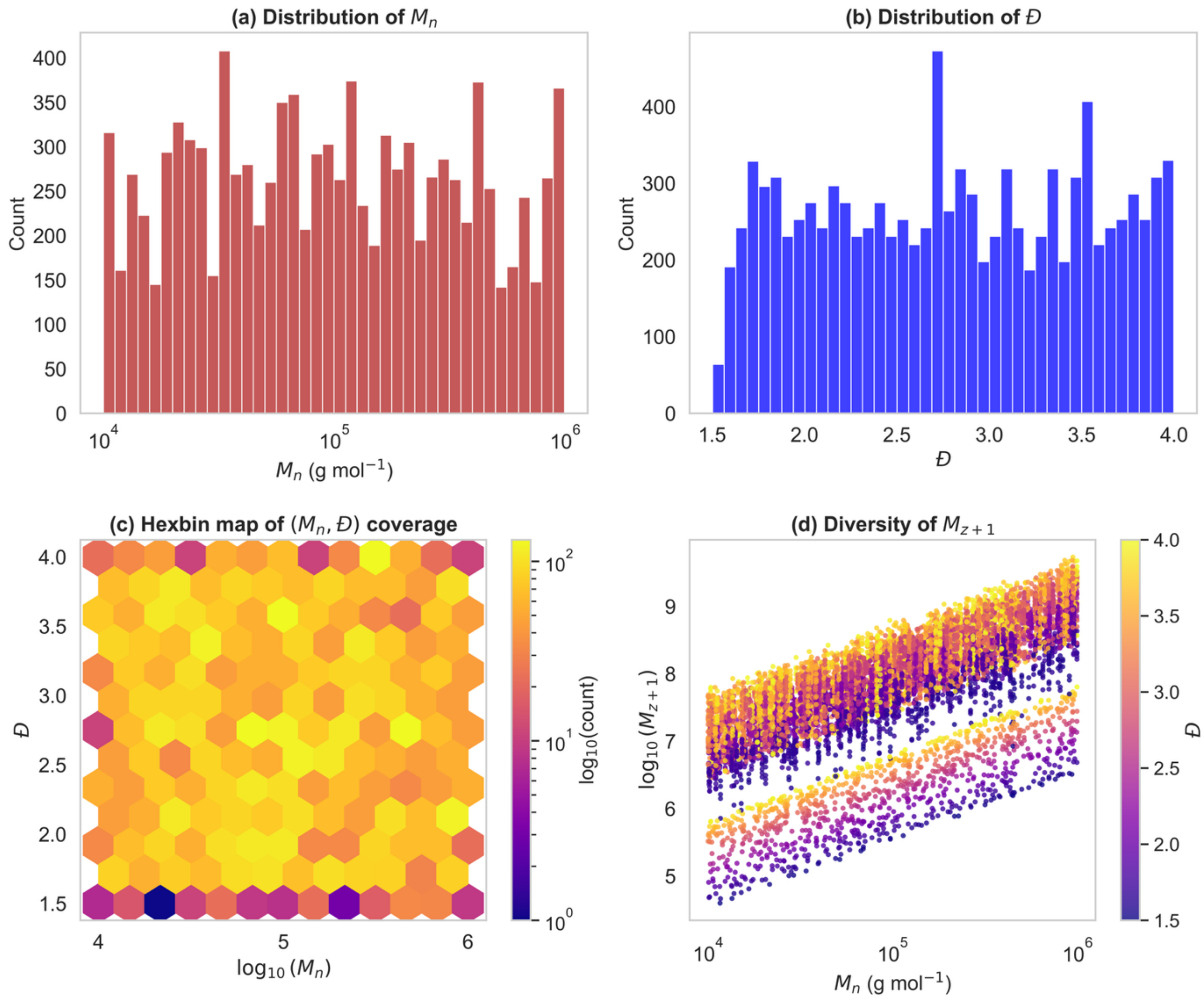

***Fig 2** - **Statistical structure of the synthetic dataset of 10 000 polymers.** (a) Distribution of number-average molar masses ($M_n$). (b) Distribution of dispersities* (Đ)*. (c) Hexbin map showing the coverage of the* ($M_n$, Đ) *domain. (d) Diversity of higher-order moments illustrated through* $\log_{10}(M_{z+1})$ *as a function of* $M_n$.

PolySet embedding

To evaluate whether PolySet preserves the distributional variability that is lost in conventional descriptors, we consider a subset of polymers sharing identical $M_n$~ $10^6$ g mol$^{-1}$ and Đ ~ 3, but exhibiting distinct MWDs (**Fig. 3a**). Although these polymers are indistinguishable from the perspective of the scalar pair ($M_n$, Đ), they differ markedly in the weight of their high-molecular-mass tail. We quantify this tail sensitivity through the (z+1)-moment. We use $\log_{10}(M_{z+1})$ as a colour annotation because this higher-order moment increases monotonically with tail heaviness and provides a robust, physically meaningful scalar to track distributional variability.

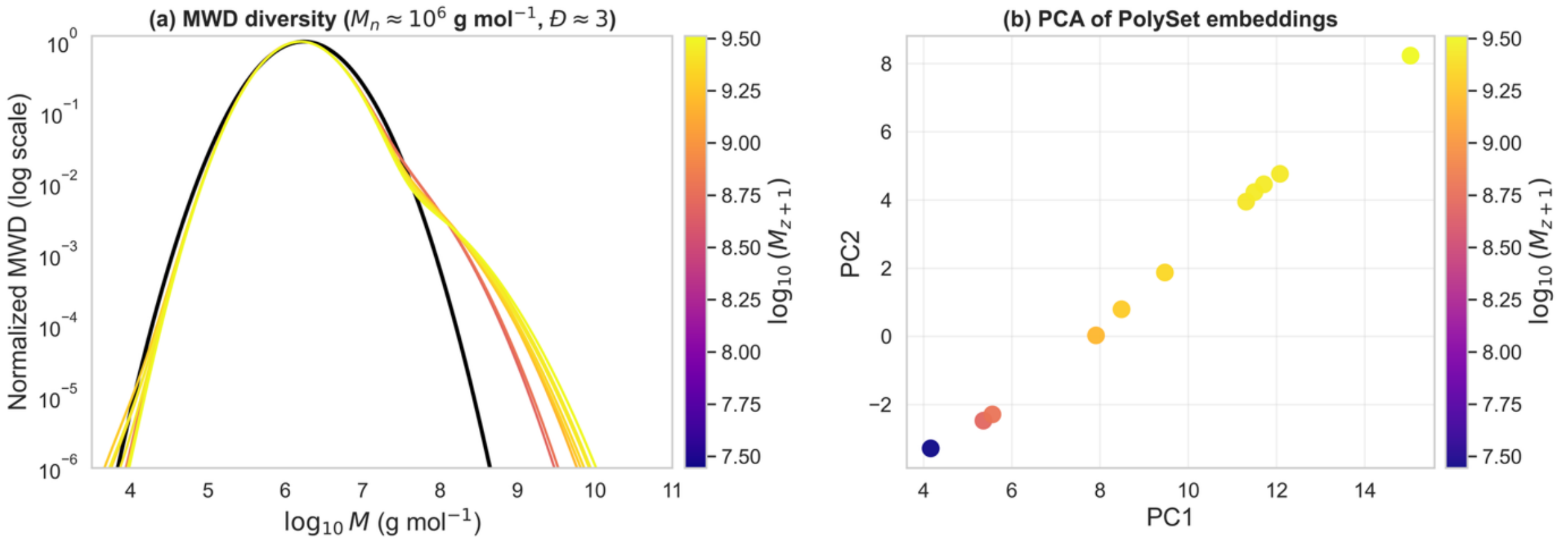

***Fig 3 - PolySet resolves distributional variability invisible to ($M_n$, Đ)***. *(a) MWD of polymers sharing identical $M_n$~$10^6$ g.mol$^{-1}$ and Đ~3, colour-coded by $\log_{10}(M_{z+1})$. (b) PCA of PolySet embeddings for the same polymer subset, coloured by $\log_{10}(M_{z+1})$. The ordered manifold reflects the underlying variation in $M_{z+1}$, demonstrating that PolySet captures distributional features indistinguishable from ($M_n$, Đ) alone.*

When encoded with the PolySet representation and projected *via* PCA (**Fig. 3b**), the embeddings of these iso-($M_n$, Đ) polymers form a smooth, low-dimensional manifold whose position correlates continuously with $\log_{10}(M_{z+1})$. This demonstrates that PolySet retains and organises the information contained in the MWD, enabling the latent space to reflect physically relevant differences between polymers that otherwise share identical first moments. By contrast, a conventional monodisperse or single-chain representation cannot be meaningfully compared here: it encodes only the repeat unit and a single hypothetical chain length, resulting in identical embeddings for all polymers in this subset and therefore no capacity to distinguish distributional variability. PolySet thus provides a distribution-aware embedding that is intrinsically sensitive to the stochastic nature of polymeric matter.

<u>PolySet improves ML stability and predictive performance</u>

To assess whether access to distributional information improves learnability, we trained a neural network model to predict $M_{z+1}$ from either (i) a conventional representation consisting of the repeat-unit motif augmented with scalaires ($M_n$, Đ), and (ii) the PolySet embedding (see Methods). Consistent with the loss of information in conventional representation, the training curves display slow and noisy convergence ($R^2$=0.484), and the resulting predictions exhibit substantial scatter (**Figs 4 a,b**). In contrast, PolySet provides the model with a distribution-resolved embedding that retains the polymer's intrinsic variability. This yields markedly improved training dynamics: the loss decays smoothly and rapidly, and the predicted $M_{z+1}$ values align closely with the ground truth, achieving $R^2$=0.998 (**Fig 4b**). The error distribution collapses dramatically, with the SMAPE reduced by more than an order of magnitude with PolySet compared to the conventional representation (**Fig 4c**). Similar distribution-resolved trends are observed for the $M_z$ moment (**Fig S2**), confirming that PolySet captures higher-order tail sensitivity beyond $M_{z+1}$.

These results demonstrate that restoring the ensemble nature of polymers within ML does not merely refine prediction—it fundamentally improves the model's ability to learn distribution-sensitive properties.

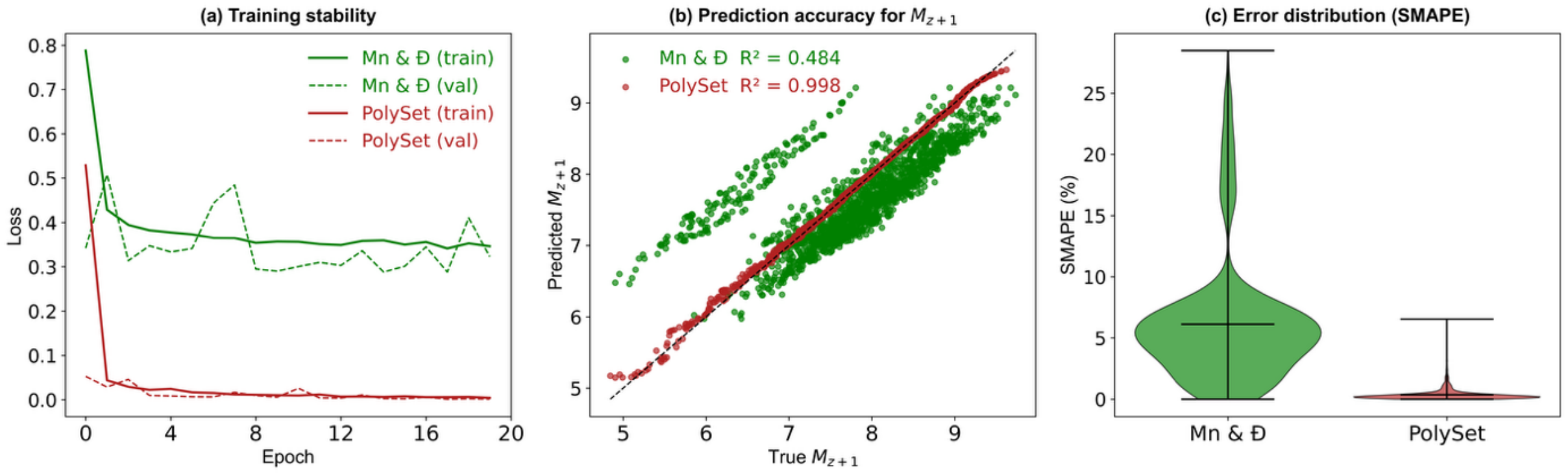

***Fig 4 - PolySet enables stable learning and accurate prediction of distribution-sensitive polymer properties.*** *(a) Learning curves: baseline vs PolySet. (b) Prediction accuracy for* $M_{z+1}$ *(c) Error distributions (SMAPE).*

## Discussion

The results presented here establish that polymer representations used in current ML models lack the fundamental ingredient that defines polymeric matter: its intrinsic stochasticity — the fact that every polymer is not a single chain, but a statistical ensemble of chains drawn from a distribution of lengths. By treating a polymer as a single hypothetical chain, conventional descriptors inevitably collapse polymers that differ only in their MWD, even when such differences strongly influence rheology, diffusivity, or viscoelasticity. PolySet resolves this limitation by reinstating the stochastic ensemble nature of polymers within the representation itself. Through a finite, distribution-aware set of weighted chains, PolySet provides a physically grounded embedding that preserves the variability encoded in the MWD. This ensemble viewpoint improves learnability. Properties that are classically difficult to predict—because they depend on the high-molecular-weight tail—become substantially easier to learn when the model receives a distribution-resolved embedding. The resulting gains in stability, accuracy, and error collapse demonstrate that the failure of previous approaches is not primarily architectural but representational: the relevant information was simply absent from the input space.

The present work focuses on homopolymers, but the conceptual framework naturally extends to more complex architectures. Random copolymers can be represented as distributions over both chain length and composition; block copolymers as distributions over block lengths; grafted, hyperbranched, or star polymers as hierarchical distributions over architectural motifs. More broadly, PolySet suggests that polymer datasets should evolve beyond scalar metadata toward distribution-level characterisation, aligning polymer informatics with the ensemble-based descriptions long used in polymer physics.

Finally, by demonstrating that ML performance improves dramatically once the representation captures the true stochastic nature of polymers, this work highlights a broader principle: the bottleneck in polymer informatics lies not in the models but in the representations. PolySet provides a minimal, physically meaningful step toward closing this gap. We anticipate that adopting distribution-aware representations will enable more accurate modelling, guide experimental design, and accelerate the development of polymer datasets that reflect the inherent statistical nature of macromolecular materials.

## Methods

PolySet Framework

PolySet represents a polymer explicitly as a finite stochastic ensemble of chains (Eq. 1):

$$P = \{(S_i, w_i)\}_{i=1}^{N} \qquad \text{(Eq. 1)}$$

Where $S_i$ denotes an individual chain in the material and $w_i$ its normalized statistical weight. Each chain is characterized by a molar mass $M_i$ and a degree of polymerization ($X_i$) (Eq. 2):

$$X_i = \frac{M_i}{M_0} \qquad \text{(Eq. 2)}$$

with $M_0$ the monomer molar mass. The weights satisfy (Eq. 3):

$$\sum_{i=1}^{N} w_i = 1 \qquad \text{(Eq. 3)}$$

Molar-mass moments and dispersity

For a homopolymer, the standard structural descriptors are the number-average molar mass ($M_n$), and the weight-average molar mass ($M_w$), defined by (Eq. 4) and (Eq. 5) respectively. The ratio (Đ) defines the dispersity, which quantifies the breadth of the MWD (Eq. 6).

$$M_n = \sum_{i=1}^{N} w_i M_i \qquad \text{(Eq. 4)}$$

$$M_w = \frac{\sum_{i=1}^{N} w_i M_i^2}{\sum_{i=1}^{N} w_i M_i} \qquad \text{(Eq. 5)}$$

$$Đ = \frac{M_w}{M_n} \qquad \text{(Eq. 6)}$$

To probe the sensitivity of polymer properties to the heavy tail of the molecular-weight distribution, we additionally consider the higher-order moments (Eq. 7)

$$M_z = \frac{\sum_{i=1}^{\mathrm{N}} w_i M_i^3}{\sum_{i=1}^{\mathrm{N}} w_i M_i^2}, \qquad M_{z+1} = \frac{\sum_{i=1}^{\mathrm{N}} w_i M_i^4}{\sum_{i=1}^{\mathrm{N}} w_i M_i^3} \qquad \text{(Eq. 7)}$$

These quantities increasingly emphasize the high-mass end of the distribution. Full derivations are provided in the Supplementary Information.

Sampling a finite chain ensemble

Given target values ($M_n$, Đ), we construct a continuous molecular-weight distribution $p(M)$. Any unimodal right-skewed distribution compatible with ($M_n$, Đ) may be used. In this work we employ a lognormal parameterized to match the prescribed moments. A finite set of chain masses $\{M_i\}_{i=1}^{N}$ is then drawn from *p(M)*, and each chain is assigned a weight (Eq. 8)

$$w_i = \frac{p(M_i)}{\sum_{i=1}^{\mathrm{N}} p(M_i)} \qquad \text{(Eq. 8)}$$

Distribution-aware embedding

Each chain $S_i$ is encoded independently into a vector representation $f(S_i)$. To isolate the effect of the distribution rather than chemical complexity, we employ a minimal sequence encoder with reduced structural bias. The polymer-level embedding is the probability-weighted aggregation (Eq. 9)

$$F_{\mathrm{P}} = \sum_{i=1}^{\mathrm{N}} w_i\, f(S_i) \qquad \text{(Eq. 9)}$$

This vector constitutes the distribution-aware representation supplied to downstream machine-learning models. It carries explicit information about chain-length variability and preserves higher-order molar-mass moments that are inaccessible to scalar descriptors.

Prediction model and training protocol

To evaluate the usefulness of PolySet embeddings, we train a regression model to predict tail-sensitive targets such as $M_{z+1}$. The predictive model is a lightweight feed-forward network that takes $F_P$ as input. The architecture is intentionally simple: no convolutions or attention mechanisms are used, ensuring that performance differences arise from the representation rather than the learning algorithm. All scalar variables are standardized, PolySet embeddings are provided as produced by the encoder, and the dataset is split into training (70%), validation (15%), and test (15%). Training is performed with the Adam optimizer and mean-squared-error loss. Full architectural details, hyperparameters, learning curves, and ablation studies are available in Zenodo.[27]

## Conclusion

This work demonstrates that the primary limitation of current polymer machine-learning models is not architectural but representational. By collapsing a polymer's inherently stochastic ensemble into a single deterministic chain, conventional descriptors eliminate the very variability that governs its physical behaviour. PolySet resolves this long-standing inconsistency by embedding polymers as finite stochastic ensembles whose distributional features remain accessible to learning algorithms.

The resulting gains are substantial: distributionally distinct polymers become separable in latent space; training becomes stable; and prediction of tail-sensitive properties such as $M_{z+1}$ reaches near-ideal accuracy. More fundamentally, PolySet shows that polymer informatics must embrace stochasticity to faithfully model macromolecular materials.

Although demonstrated here for homopolymers, the framework generalises naturally to random and block copolymers, gradient architectures, and grafted or hyperbranched systems, each defined by its own hierarchy of distributions. By restoring the ensemble ontology that has always underpinned polymer physics, PolySet provides a physically grounded and extensible representation capable of supporting the next generation of data-driven polymer design.

## Data availability

The full synthetic dataset, together with scripts for reproduction, is available in Zenodo.[27]

## Code availability

A Python package implementing PolySet for academic use is available on PyPI (polysetlib) and GitHub (https://github.com/kFERJI/PolySet).

## References

1 Staudinger, H. Über Polymerisation. *Berichte der deutschen chemischen Gesellschaft (A and B Series)* **53**, 1073-1085 (1920).

2 Flory, P. J. Viscosities of Linear Polyesters. An Exact Relationship between Viscosity and Chain Length. *J. Am. Chem. Soc.* **62**, 1057-1070 (1940).

3 Berry, G. C. & Fox, T. G. The Viscosity of Polymers and their Concentrated Solutions. *Advances in Polymer Science* **5**, 261–357 (1968).

4 Fox Jr, T. G. & Flory, P. J. Second-order transition temperatures and related properties of polystyrene. I. Influence of molecular weight. *Journal of Applied Physics* **21**, 581-591 (1950).

5 Keith, H. D. & Padden Jr, F. A phenomenological theory of spherulitic crystallization. *Journal of Applied Physics* **34**, 2409-2421 (1963).

6 Van Holde, K. E. & Williams, J. W. Study of the viscoelastic behavior and molecular weight distribution of polyisobutylene. *J. Polym. Sci.* **11**, 243-268 (1953).

7 Zimm, B. H. Dynamics of polymer molecules in dilute solution: viscoelasticity, flow birefringence and dielectric loss. *The journal of chemical physics* **24**, 269-278 (1956).

8 Schulz, G. V. Über die Kinetik der Kettenpolymerisationen. V. *Zeitschrift für Physikalische Chemie* **43B**, 25-46 (1939).

9 Flory, P. J. Molecular size distribution in linear condensation polymers1. *J. Am. Chem. Soc.* **58**, 1877-1885 (1936).

10 Audus, D. J. & de Pablo, J. J. Polymer Informatics: Opportunities and Challenges. *ACS Macro Lett.* **6**, 1078-1082 (2017).

11 Ge, W., De Silva, R., Fan, Y., Sisson, S. A. & Stenzel, M. H. Machine Learning in Polymer Research. *Adv. Mater.* **37**, 2413695 (2025).

12 Ferji, K. Basic concepts and tools of artificial intelligence in polymer science. *Polym. Chem.* **16**, 2457-2470 (2025).

13 Chen, L. *et al.* Polymer informatics: Current status and critical next steps. *Mater. Sci. Eng. R Rep.* **144**, 100595 (2021).

14 Doan Tran, H. *et al.* Machine-learning predictions of polymer properties with Polymer Genome. *Journal of Applied Physics* **128** (2020).

15 Wang, F. *et al.* in *Proceedings of the 33rd ACM International Conference on Information and Knowledge Management.* 2336-2346.

16 Zeng, M. *et al.* Graph convolutional neural networks for polymers property prediction. *arXiv preprint arXiv:1811.06231* (2018). https://doi.org:10.48550/arXiv.1811.06231

17 Whitfield, R. *et al.* Tailoring polymer dispersity and shape of molecular weight distributions: methods and applications. *Chem. Sci.* **10**, 8724-8734 (2019).

18 Weininger, D. SMILES, a chemical language and information system. 1. Introduction to methodology and encoding rules. *Journal of Chemical Information and Computer Sciences* **28**, 31-36 (1988).

19 Lin, T.-S. *et al.* BigSMILES: A Structurally-Based Line Notation for Describing Macromolecules. *ACS Central Science* **5**, 1523-1531 (2019).

20 Schneider, L., Walsh, D., Olsen, B. & de Pablo, J. Generative BigSMILES: an extension for polymer informatics, computer simulations & ML/AI. *Digital Discovery* **3**, 51-61 (2024).

21 Krenn, M. *et al.* SELFIES and the future of molecular string representations. *Patterns* **3**, 100588 (2022).

22 Zhou, J., Yang, Y., Mroz, A. M. & Jelfs, K. E. PolyCL: contrastive learning for polymer representation learning via explicit and implicit augmentations. *Digital Discovery* **4**, 149-160 (2025).

23 Kunchapu, S. & Jablonka, K. M. PolyMetriX: an ecosystem for digital polymer chemistry. *npj Comput. Mater.* **11**, 312 (2025).

24 Jain, A., Gurnani, R., Rajan, A., Qi, H. J. & Ramprasad, R. A physics-enforced neural network to predict polymer melt viscosity. *npj Comput. Mater.* **11**, 42 (2025).

25 Kuenneth, C. & Ramprasad, R. polyBERT: a chemical language model to enable fully machine-driven ultrafast polymer informatics. *Nat Commun* **14**, 4099 (2023).

26 Ishii, M., Ito, T., Sado, H. & Kuwajima, I. NIMS polymer database PoLyInfo (I): an overarching view of half a million data points. *Science and Technology of Advanced Materials: Methods* **4**, 2354649 (2024).

27 FERJI, K. PolySet: Statistical Ensemble Polymer Embeddings Dataset for Machine Learning. *Zenodo* (2025). https://doi.org:10.5281/zenodo.17861022

# Supporting Information

## PolySet: Restoring the Statistical Ensemble Nature of Polymers for Machine Learning

Khalid FERJI[1]

LCPM, Université de Lorraine, UMR 7375 Nancy – France

Khalid.ferji@univ-lorraine.fr

https://orcid.org/0000-0003-3073-9722

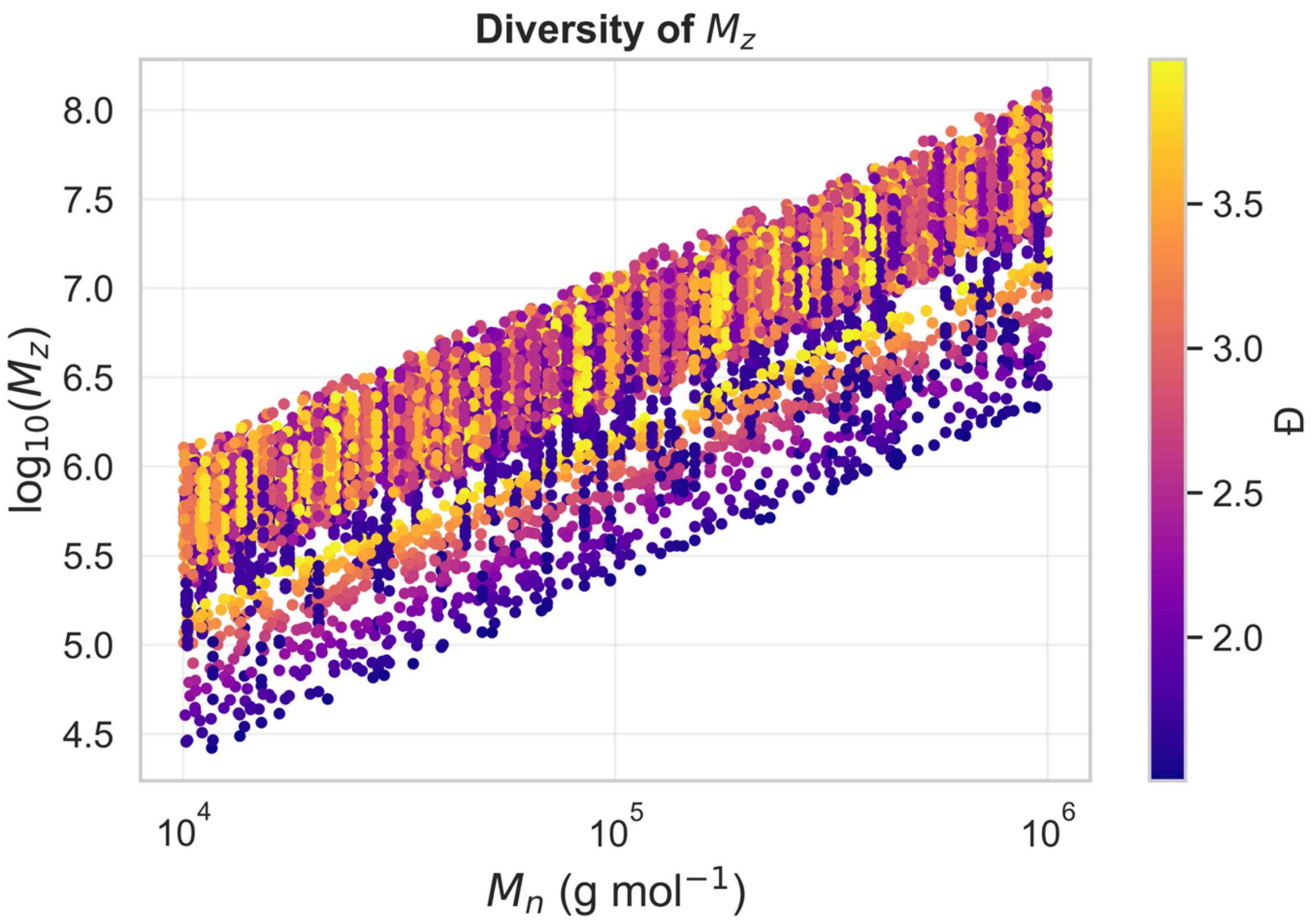

**Fig S1- Diversity of higher-order molar-mass moment ($M_z$) across the synthetic polymer dataset.**

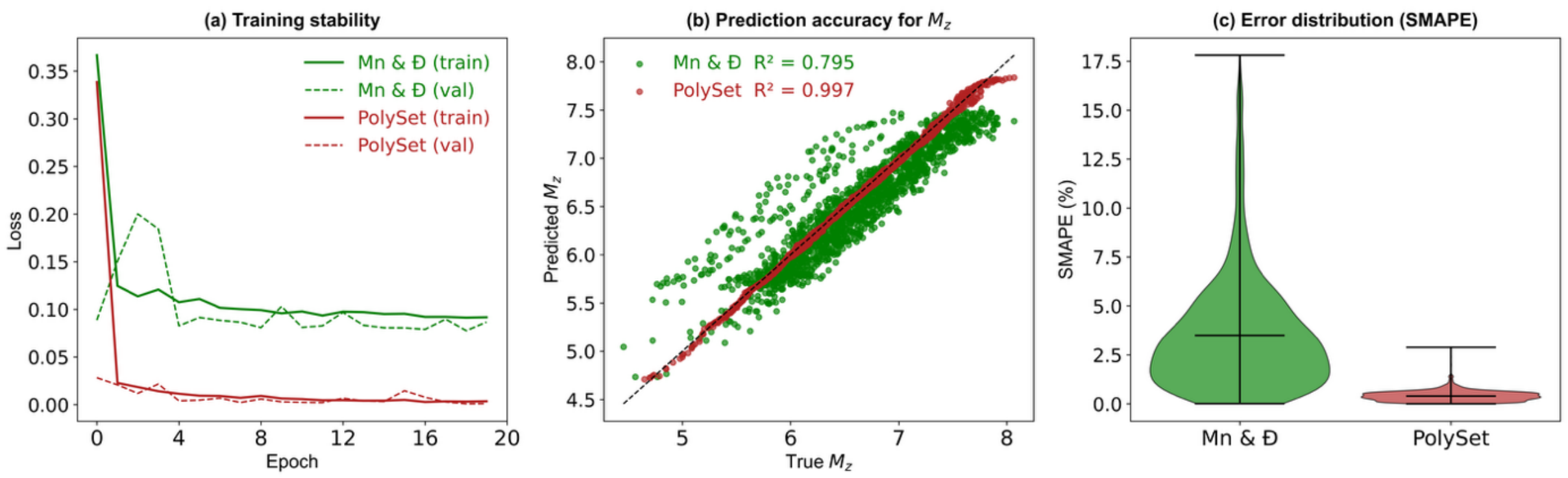

**Fig S2 – *PolySet enables stable learning and accurate prediction of distribution-sensitive polymer properties.*** *(a) Learning curves: baseline vs PolySet. (b) Prediction accuracy for $M_z$ (c) Error distributions (SMAPE).*